\documentclass[letterpaper]{article}
\PassOptionsToPackage{table}{xcolor}
\usepackage[preprint]{aaai2027}
\usepackage[hyphens]{url}
\usepackage{graphicx}
\usepackage{natbib}
\usepackage{caption}
\usepackage{amsmath}
\usepackage{amssymb}
\usepackage{booktabs}
\usepackage{siunitx}
\usepackage{algorithm}
\usepackage{algpseudocode}

\definecolor{bestcell}{gray}{0.92}
\renewcommand{\epsilon}{\varepsilon}
\newcommand{\tblval}[1]{\num{#1}}
\newcommand{\tblbest}[1]{\cellcolor{bestcell}\textbf{\num{#1}}}

\title{Staged Depth Training: A Representation Curriculum\\for PINNs}
\author{
    Kejia Zhang\equalcontrib,
    Youran Sun\equalcontrib,
    Haizhao Yang\corresponding
}
\affiliations{
    University of Maryland, College Park\\
    hzyang@umd.edu
}

\begin{document}

\maketitle

\begin{abstract}
Representation quality is a central determinant of PINNs' performance, yet standard training leaves representations to emerge implicitly while fitting the final solution.
We introduce \textbf{representation curriculum}, an ordered process in which representations are explicitly learned, transferred independently of their predictors, and progressively refined.
We realize it with Staged Depth Training (SDT), which trains a shallow prefix under a temporary physics-informed head, discards the head, and freezes the learned prefix while adding depth, without equation-specific encodings or changes to the final architecture. 
Across the 20 default forward problems in PINNacle with three backbones, SDT improves 40 of 59 equal-budget problem--backbone cells by at least 5\% and remains within that band in the rest, with a 32.8\% geometric-mean error reduction on a PirateNet-style backbone.
Mechanistic ablations suggest that the gain is not explained by optimizer restarts or shallow warm-starting alone.
Representation visualizations and hyperparameter-basin analyses provide diagnostic evidence on representation geometry and local sensitivity to shared hyperparameters.
On Poisson--Boltzmann 2D, SDT also more than doubles the fitted depth-scaling exponent for both backbones.
These results support representation curriculum as a promising training strategy for improving PINNs while preserving the deployed architecture and inference cost.
\end{abstract}

\section{Introduction}
\label{sec:introduction}

Representations strongly influence PINN training, yet hidden representations are typically learned as by-products of fitting the final solution.
A standard PINN \cite{Raissi2019PINNs} is trained as a monolithic predictor whose hidden states are optimized indirectly through its output, and the representation itself is rarely treated as an object that can be supervised, retained, and reused.
How can a PINN learn its internal representation explicitly, rather than leaving it as a by-product of end-to-end solution fitting?

Existing coordinate encodings demonstrate the importance of representation: engineered feature spaces \cite{Wang2021EigenvectorBias,Zeng2024RBFPINN,Huang2024HashEncoding,Fazliani2025SAFENET} expose structures that raw coordinates make difficult to learn, but their effectiveness depends on human choices about feature families, frequencies, scales, and resolutions, and even trainable encodings prescribe the basic form of the feature space.
\emph{A natural representation should be learned from the PDE, rather than designed separately for every equation.}

We elevate the hidden representation to an explicit learning object that should receive stage-local physics supervision through a temporary predictor, remain useful after that predictor is removed, and continue to develop during training.
We define a representation curriculum as an ordered learning process in which physics-trained hidden representations are learned through stage-specific heads, transferred independently of those heads, and progressively refined; to our knowledge, this is the first representation curriculum for PINNs.
At each stage, a stage-specific predictor connects the current representation to the solution, allowing the original physics objective to train the current prefix through a shortened credit path.

End-to-end training couples the entire representation hierarchy from the first update, so early representations must continually adapt to downstream mappings that are themselves evolving from random initialization.
Auxiliary heads \cite{Lee2015DeeplySupervised} strengthen intermediate supervision, yet every representation still co-adapts inside the complete network, so later layers never learn from a representation that has first been formed as a useful feature space.

Progressive training strategies look similar at the level of their schedules, but none satisfies all three requirements.
Progressive widening \cite{Huang2022PINNup,Xu2023GreedyPDE} expands the approximation space of the current predictor, and layerwise training \cite{Krishnanunni2025Layerwise} keeps the learned features coupled to the growing predictor.
The distinction is especially direct for multigrade learning \cite{Xu2026MGDL,XuZeng2023MGDL} and stacked PINNs \cite{Howard2025Stacked}, which carry a predictor output forward rather than a hidden representation.

\begin{figure*}[ht]
    \centering
    \includegraphics[width=0.97\linewidth,trim=40 50 40 40,clip]{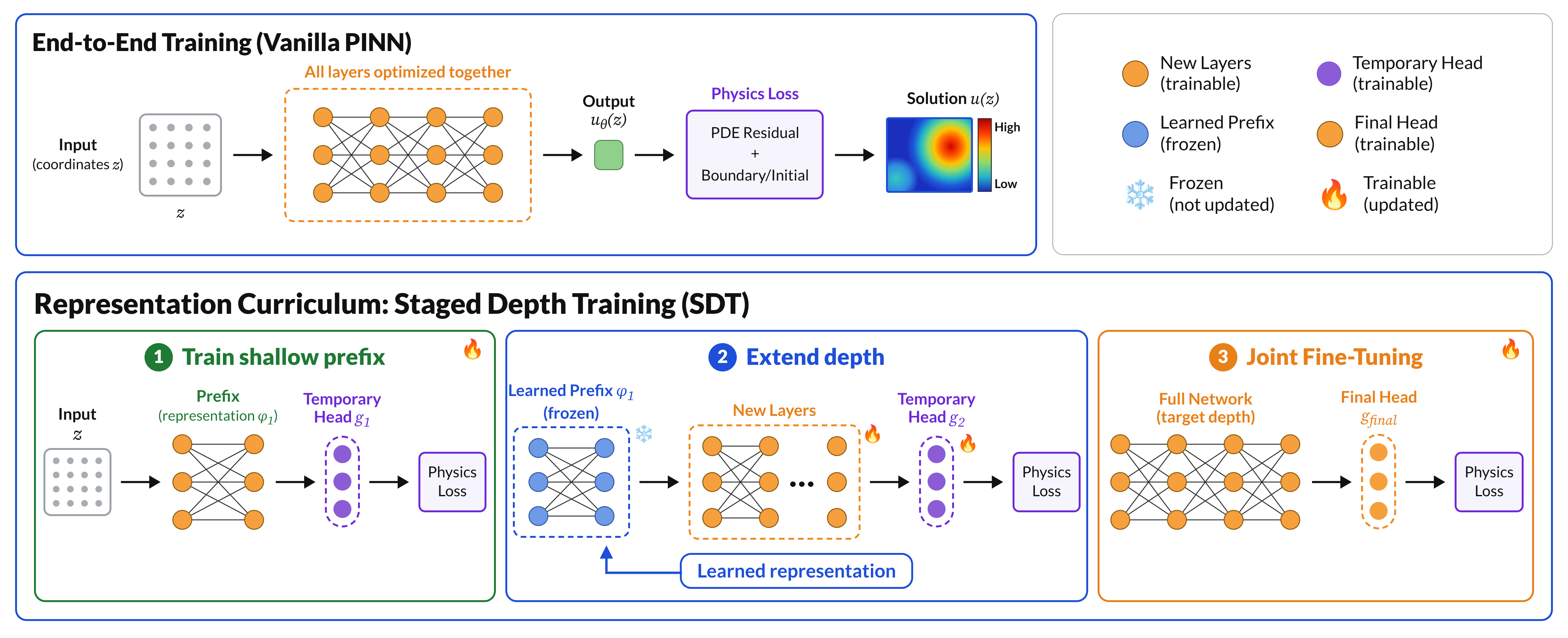}
    \caption{Overview of representation curriculum for PINNs. End-to-end training optimizes the whole hierarchy through the final output, whereas SDT learns a shallow representation under a temporary physics-informed head, freezes and transfers it while new layers train, and finally fine-tunes the full network. Every temporary head is discarded, so the deployed model matches the end-to-end baseline.}
    \label{fig:sdt-overview}
\end{figure*}

Together, these considerations sharply constrain the design of the representation curriculum.
Stage-local physics supervision of the current hidden representation is implemented with a temporary physics-informed head after a shallow network prefix, so that the complete PINN objective trains the current shallow network.
Once the stage ends, the head is discarded and the learned hidden prefix is retained.
To keep this representation stable while deeper features are formed, we temporarily freeze the prefix, append new layers and a fresh temporary head, and train the new layers on top of the existing representation.
Repeating this process learns progressively deeper representations until the target depth is reached, after which we unfreeze all layers and jointly fine-tune the complete PINN.
We call this procedure Staged Depth Training (SDT), summarized in Figure~\ref{fig:sdt-overview}.

SDT is architecturally compatible with backbones that expose exact hidden prefixes and uses no equation-specific encoding or modified physics objective; every temporary head is removed after training, leaving the same final architecture, parameter count, and inference graph as end-to-end training.
Across all 20 default forward problems in PINNacle \cite{ZhongkaiHao2024PINNacle} and three backbone families, SDT achieves lower relative $L_2$ error in 53 of 58 non-tied equal-budget comparisons and reduces geometric-mean error by 28.2\% to 32.8\% per backbone.
It also outperforms optimizer-restart and shallow warm-start controls in all four mechanistic settings, and a t-SNE embedding of its hidden representation shows 33.1\% less local solution-value mixing on a multiscale PDE.
Earlier stages train smaller networks, reducing the architecture-level training MAC proxy by 27.8\% on average while preserving identical inference cost.

Our contributions are fourfold:
\begin{itemize}
    \item We introduce representation curriculum as a training perspective for PINNs, making hidden representations stage-trained, transferable, and progressively refined learning objects.
    \item We develop SDT, a backbone-agnostic realization of representation curriculum based on temporary physics-informed heads, hidden-prefix transfer, and staged depth extension.
    \item We provide empirical evidence across the 20 default PINNacle forward problems and three backbone families: SDT improves 40 of 59 equal-budget cells by at least 5\%, with the rest within that descriptive band.
    Four-arm controls point to protected representation transfer as a key ingredient, and representation visualization and hyperparameter-basin experiments add complementary diagnostics on mechanism and local sensitivity.
    \item On Poisson--Boltzmann 2D, we show that SDT raises the fitted depth-scaling exponent $\gamma$ from 0.29 to 0.74 for MLP and from 0.31 to 0.78 for a PirateNet-style backbone.
\end{itemize}

\section{Related Work}
\label{sec:related-work}

Prior work separates by the object it learns: a prescribed representation, an enlarged predictor, or a sequence of solution corrections.
Appendix~\ref{app:positioning} gives a focused comparison with the closest neighboring methods.

\paragraph{Prescribed representations.}
Multiscale Fourier features \cite{Wang2021EigenvectorBias}, radial-basis embeddings \cite{Zeng2024RBFPINN}, multiresolution hash encodings \cite{Huang2024HashEncoding}, and SAFE-NET \cite{Fazliani2025SAFENET} supply feature spaces exposing frequencies, local structure, and multiple scales, highlighting the importance of representation while leaving the feature family, scales, and geometric priors to human design.
LatentPINNs \cite{Taufik2025LatentPINNs} learn latent variables encoding PDE parameters, so the learned object represents the task supplied to the solver rather than the network's own hidden features.

\paragraph{Enlarged predictors.}
Greedy layerwise pretraining \cite{Hinton2006DBN,Bengio2006GreedyLayerwise} showed that intermediate representations can be built sequentially before global refinement.
The closest non-PINN precedents grow large networks while reusing weights: function-preserving growth and progressive stacking of Transformer depth \cite{Chen2016Net2Net,Gong2019StackingBERT,Shen2022StagedTraining}, progressively larger trained subnetworks \cite{Panigrahi2025RAPTR}, and layer growth combined with curricula or freezing rules \cite{Singh2026CGLS,Erdogan2025LayerLock}.
These show that the training trajectory is a design variable at a fixed target architecture, but what crosses a boundary stays part of the predictor being grown.
Their neural-PDE counterparts enlarge capacity: PINNup splits neurons \cite{Huang2022PINNup}, greedy solvers append neurons or dictionary elements \cite{Xu2023GreedyPDE}, and multilevel-in-width training builds a hierarchy of wider networks \cite{Ponce2023Multilevel}.
Each stage expands the function class of the current predictor rather than separating a hidden representation from the predictor that produced it.
Depth-oriented methods come closer: PirateNets \cite{SifanWang2024PirateNets} instantiate every layer from the start and gradually activate deeper blocks, so the whole hierarchy co-adapts, while layerwise PINN training \cite{Krishnanunni2025Layerwise} freezes trained components but inherits its prediction head, keeping features coupled to the predictor under construction.

\paragraph{Accumulated predictions.}
Multi-grade deep learning \cite{Xu2026MGDL,XuZeng2023MGDL} freezes earlier grades and fits each new grade to the error the accumulated solution leaves behind, formalized by recent analyses as hierarchical refinement \cite{Zhang2026MultigradeApprox,Zhang2026GeometricLayerwise}; stacked PINNs \cite{Howard2025Stacked} pass one network's output into the next.
These stage the physics training, but a prediction crosses each boundary and remains part of the final approximation, so the hidden representation stays coupled to the predictor formed alongside it.

Deep supervision \cite{Lee2015DeeplySupervised} adds auxiliary heads but keeps every representation co-adapting inside one jointly optimized network.
No family combines the three properties a representation curriculum requires: stage-local physics supervision through a temporary head, predictor-independent representation transfer, and progressive refinement inside a single network.

\section{Method}
\label{sec:method}

\subsection{Physics-Informed Objective}
\label{sec:pinn-objective}

Write $\boldsymbol{z}=\boldsymbol{x}\in\Omega$ for stationary problems and $\boldsymbol{z}=(\boldsymbol{x},t)\in\Omega\times[0,T_f]$ otherwise, and let $\mathcal{Z}$ denote the resulting domain.
The target solution $\boldsymbol{u}$ satisfies $J_r$ interior equations $\mathcal{N}_j[\boldsymbol{u}]=\boldsymbol{0}$ on $\mathcal{Z}$, $J_b$ boundary conditions $\mathcal{B}_j[\boldsymbol{u}]=\boldsymbol{g}_j$ on $\partial\Omega$ (times $[0,T_f]$ when time is present), and, for time-dependent problems, $J_i$ initial conditions $\mathcal{I}_j[\boldsymbol{u}](\cdot,0)=\boldsymbol{u}_{0,j}$ on $\Omega$.
A physics-informed neural network $\boldsymbol{u}_{\theta}$ approximates $\boldsymbol{u}$ and obtains every required derivative by automatic differentiation~\cite{Raissi2019PINNs}.
Collecting these constraints, their collocation sets, and their right-hand sides into triples $(\mathcal{C}_j,\mathcal{X}_j,\boldsymbol{c}_j)$, we minimize the unit-weighted objective
\begin{equation}
    \mathcal{L}(\theta)
    = \sum_{j}
      \frac{1}{|\mathcal{X}_j|}\sum_{\boldsymbol{z}\in\mathcal{X}_j}
      \bigl\|\mathcal{C}_j[\boldsymbol{u}_{\theta}](\boldsymbol{z})-\boldsymbol{c}_j(\boldsymbol{z})\bigr\|_2^2 .
    \label{eq:pinn-loss}
\end{equation}
The experiments use the operators, domains, and collocation sets supplied by PINNacle~\cite{ZhongkaiHao2024PINNacle}, and reference solutions are never added as interior targets.
SDT leaves Eq.~\eqref{eq:pinn-loss} untouched: every stage optimizes the same objective on the same collocation sets, and only the trainable parameter set and the network depth change from stage to stage.

\subsection{Representation Curriculum and Staged Depth Training}
\label{sec:staged-training}

Throughout, \emph{vanilla} (end-to-end) training means optimizing the complete final-depth network from initialization for the entire budget, with no depth expansion, parameter freezing, or stage transition.

Let $K\geq2$ and fix cumulative hidden depths $D_1<D_2<\cdots<D_K=D$.
We call the depth-$D_k$ hidden prefix, together with the feature map it induces on the input coordinates, the \emph{representation at level $k$}; the head that reads it is a stage-specific predictor and is not part of the representation.
SDT traverses the $K$ levels and closes with an all-parameter fine-tuning stage, giving $K+1$ optimization stages.
Stage one trains a depth-$D_1$ network together with an output head for $T_1$ updates.
At growth stage $k=2,\ldots,K$, a depth-$D_k$ network and a fresh head are initialized, the learned level-$(k{-}1)$ prefix is copied in and frozen, the preceding head is discarded, and only the newly added layers and the new head are trained for $T_k$ updates.
The final stage unfreezes the complete depth-$D_K$ network and jointly fine-tunes all its parameters for $T_{K+1}$ updates.
Thus $K$ counts depth levels rather than optimization stages: $(D_1,D_2)=(3,6)$ gives three stages, whereas $(D_1,D_2,D_3,D_4)=(3,6,9,12)$ gives five.

Three elements realize the three properties of a representation curriculum.
The temporary head gives the current representation stage-local physics supervision through a shortened credit path, so it is trained by the PDE itself rather than by a surrogate target.
Discarding that head and freezing the prefix transfers a representation instead of a predictor.
Repeating the construction refines the representation one level at a time until the target depth is reached.

We parameterize the stage lengths by positive fractions $f_1,\ldots,f_{K+1}$ summing to one, with $T_k=\lfloor f_kT\rfloor$ for $k\leq K$ and the exact remainder assigned to $T_{K+1}$.
Each stage holds its learning rate $\eta_k$ constant, and a fresh optimizer is constructed at every transition, so optimizer moments never cross a stage boundary.
Algorithm~\ref{alg:staged-depth} summarizes the procedure.

\begin{algorithm}[t]
\caption{Staged Depth Training with $K$ depth levels}
\label{alg:staged-depth}
\begin{algorithmic}[1]
\Require Depths $D_1<\cdots<D_K$, updates $T$, fractions $(f_1,\ldots,f_{K+1})$, learning rates $(\eta_1,\ldots,\eta_{K+1})$
\State Set $T_k=\lfloor f_kT\rfloor$ for $k\leq K$ and $T_{K+1}=T-\sum_{k=1}^{K}T_k$
\State Initialize a depth-$D_1$ network $F^{(1)}$ and head $H^{(1)}$
\State Optimize all parameters of $H^{(1)}\circ F^{(1)}$ for $T_1$ updates at $\eta_1$
\For{$k=2,\ldots,K$}
    \State Initialize a depth-$D_k$ network $F^{(k)}$ and head $H^{(k)}$
    \State Copy the depth-$D_{k-1}$ hidden prefix from $F^{(k-1)}$ to $F^{(k)}$
    \State Freeze the copied prefix and discard $H^{(k-1)}$
    \State Optimize the new suffix and $H^{(k)}$ for $T_k$ updates at $\eta_k$
\EndFor
\State Unfreeze all parameters of $F^{(K)}$
\State Optimize $H^{(K)}\circ F^{(K)}$ jointly for $T_{K+1}$ updates at $\eta_{K+1}$
\State \Return the target-depth network $H^{(K)}\circ F^{(K)}$
\end{algorithmic}
\end{algorithm}

\subsection{Why Freezing Changes the Stage Subproblem}
\label{sec:coupling}

This subsection gives the optimization intuition behind staging; it is not a convergence guarantee.
What distinguishes physics supervision is that the hidden representation enters the loss through its coordinate derivatives and not only through its values, so how much a representation moves during training matters more the higher the differential order of the residual.

Split the network at depth $D_{k-1}$ as $\boldsymbol{u}_{\theta}=\boldsymbol{g}_{\psi}\circ\boldsymbol{\phi}_{\omega}$, where $\boldsymbol{\phi}_{\omega}:\mathbb{R}^{d}\to\mathbb{R}^{w}$ is the transferred representation and $\boldsymbol{g}_{\psi}:\mathbb{R}^{w}\to\mathbb{R}^{o}$ is the newly added suffix together with the current head.
Because the operators $\mathcal{N}_j$ differentiate with respect to $\boldsymbol{z}$, the chain rule gives
\begin{equation}
    \frac{\partial u_a}{\partial z_i}
    = \sum_{p}
      \frac{\partial g_a}{\partial h_p}
      \frac{\partial \phi_p}{\partial z_i},
    \label{eq:chain-first}
\end{equation}
and every further differentiation contracts derivatives of $\boldsymbol{g}_{\psi}$ against higher derivatives of $\boldsymbol{\phi}_{\omega}$.
A residual of differential order $m$ therefore depends on the transferred representation only through its $m$-jet
\begin{equation}
    \mathcal{J}_{\omega}^{(m)}(\boldsymbol{z})
    = \bigl(
        \boldsymbol{\phi}_{\omega},\;
        \nabla_{\boldsymbol{z}}\boldsymbol{\phi}_{\omega},\;
        \ldots,\;
        \nabla^{m}_{\boldsymbol{z}}\boldsymbol{\phi}_{\omega}
      \bigr)(\boldsymbol{z}),
    \label{eq:prefix-jet}
\end{equation}
so each stage residual is an algebraic expression in $\mathcal{J}_{\omega}^{(m)}$ and the derivatives of $\boldsymbol{g}_{\psi}$.

Under end-to-end training $\omega$ and $\psi$ move together, so the whole jet changes at every step.
A prefix trained under a pointwise regression loss also drifts, but only its values $\boldsymbol{\phi}_{\omega}(\boldsymbol{z})$ enter that loss; here the $m$ derivative levels above the values drift as well, and they enter the residual multiplied by derivatives of the suffix.
The suffix is therefore asked to satisfy a differential constraint whose coefficient fields are themselves in motion, and the number of moving coefficient fields grows with the differential order of the PDE.
This derivative-jet dependence, rather than co-adaptation as such, is the PINN-specific part of representation--optimization coupling.

Freezing removes that motion, and two properties hold throughout growth stage $k$.
First, because $\omega$ and the collocation sets are both fixed, $\mathcal{J}_{\omega}^{(m)}$ is fixed across optimizer updates on those sets, though it still varies with $\boldsymbol{z}$; every residual then becomes a differential expression in $\boldsymbol{g}_{\psi}$ alone whose coefficient fields do not move, so the suffix is optimized against a fixed learned representation rather than one that is still forming.
Second, since $\nabla_{\omega}\mathcal{L}=0$, the stage optimizes a strictly lower-dimensional problem over $\psi$.
Neither property implies that the staged optimum is better, and we claim no such thing; each states only that the stage-$k$ subproblem is stationary in a sense that end-to-end training is not.
The final stage releases $\omega$ so that representation and solver can co-adapt, but it begins from a suffix already matched to the representation beneath it rather than from an arbitrary initialization.
Section~\ref{sec:mechanism} probes this account: warm-starting from the same prefix without freezing forgoes both properties and reaches a higher final relative $L_2$ error than SDT in the tested settings.

\subsection{Design Choices}
\label{sec:design-choices}

\paragraph{Discard the head, freeze the prefix.}
A head maps the current representation to solution values at the current depth, so a head fitted at depth $D_{k-1}$ is misspecified once the representation beneath it changes; retaining it would also change what the method transfers, because the object crossing the boundary would then be a solution predictor rather than a representation.
Copying the prefix and immediately training it would likewise reinstate the coupling that staging removes: the suffix would again chase a moving $\mathcal{J}_{\omega}^{(m)}$, and the large early gradients of a freshly initialized suffix would propagate into a representation that has not yet been exploited.
Freezing gives the suffix a stationary target for the stage, and Section~\ref{sec:mechanism} shows that a prefix remains useful even when the shallow model that produced it has high standalone error.

\paragraph{Unfreeze at the end, and choose admissible depths.}
Freezing is a means rather than an end: the transferred representation was optimized under a shallower head and is not optimal for the deeper composition, so never unfreezing would leave the target network constrained by an objective it no longer faces.
Staging also requires each smaller network to be an exact hidden prefix of the next, which constrains the admissible $D_k$ per backbone; the schedule $(D_1,D_2)=(3,6)$ satisfies this for all three backbones below.

\subsection{Backbone Instantiations}
\label{sec:backbones}

Staging is defined on an ordered hidden representation, so it applies to several standard PINN backbones; all three instantiations use an elementwise hidden activation $\sigma$ and a bias-free linear output head.
For an \textbf{MLP} of depth $D$ the hidden states are $\boldsymbol{h}_{\ell}=\sigma(W_{\ell}\boldsymbol{h}_{\ell-1}+\boldsymbol{b}_{\ell})$ with $\boldsymbol{h}_0=\boldsymbol{z}$, and the level-$k$ representation consists of the first $D_{k}$ affine--activation layers.
The \textbf{ResNet} groups its hidden transformations into two-layer residual blocks after an input projection, so admissible depths must respect block boundaries; an odd hidden-layer count terminates with one additional affine--$\sigma$ transformation.
Our \textbf{PirateNet-style} backbone uses the zero-initialized adaptive residual connection of PirateNets~\cite{SifanWang2024PirateNets} with a single dense layer per block,
\begin{equation}
    \boldsymbol{h}_{\ell}
    = \boldsymbol{h}_{\ell-1}
      + \alpha_{\ell}\,\sigma\!\left(W_{\ell}\boldsymbol{h}_{\ell-1}+\boldsymbol{b}_{\ell}\right),
    \qquad \ell=1,\ldots,D,
    \label{eq:pirate-block}
\end{equation}
after $\boldsymbol{h}_0=\sigma(W_{\mathrm{in}}\boldsymbol{z}+\boldsymbol{b}_{\mathrm{in}})$, with every adaptive coefficient $\alpha_{\ell}$ initialized to zero.
The input projection and the first $D_{k-1}$ blocks form the transferred representation, each $\alpha_{\ell}$ moving with its block.
Because the newly added blocks start at $\alpha_{\ell}=0$, an expanded network initially reproduces the function computed by its transferred representation and head, so staging composes with soft deepening rather than competing with it.
Appendix~\ref{app:model-details} gives the full layer equations and exact parameter counts.

\subsection{What the Curriculum Preserves}
\label{sec:model-equivalence}

SDT changes how the hidden representation is learned, not the model returned after training.
For a fixed PDE and backbone, vanilla and staged training return exactly the same depth-$D_K$ module class, tensor shapes, parameter count, and forward graph; every temporary head is absent from the returned model, so the two arms are comparable at a fixed inference cost.
Training cost is lower, because early stages train a shallower network and growth stages backpropagate only through the trainable suffix and head.
Appendix~\ref{app:reproducibility} instantiates a dense-layer multiply--accumulate (MAC) proxy from the exact stage lengths and layer dimensions.
That proxy counts dense-layer arithmetic only and excludes the automatic-differentiation graph built for the differential operators, which is PDE-dependent and dominates for high-order residuals, so it is an architecture-level comparison rather than a prediction of wall-clock time.

\section{Experiments}
\label{sec:experiments}

\begin{table*}[t]
    \centering
    {\scriptsize
    \setlength{\tabcolsep}{1mm}
    \renewcommand{\arraystretch}{0.92}
    \begin{tabular}{lcccccc}
        \toprule
        & \multicolumn{2}{c}{MLP}
        & \multicolumn{2}{c}{ResNet}
        & \multicolumn{2}{c}{PirateNet-style} \\
        \cmidrule(lr){2-3}\cmidrule(lr){4-5}\cmidrule(lr){6-7}
        Problem
        & Vanilla & SDT (ours)
        & Vanilla & SDT (ours)
        & Vanilla & SDT (ours) \\
        \midrule
        Burgers 1D
        & \tblval{1.426e-02} & \tblbest{1.320e-02}
        & \tblval{1.289e-02} & \tblval{1.313e-02}
        & \tblval{1.360e-02} & \tblbest{1.253e-02} \\
        Burgers 2D
        & \tblval{5.236e-01} & \tblval{5.068e-01}
        & \tblval{4.915e-01} & \tblval{5.059e-01}
        & \tblval{4.795e-01} & \tblval{4.868e-01} \\
        Poisson 2D (classic)
        & \tblval{6.261e-01} & \tblval{5.968e-01}
        & \tblval{6.722e-01}$^{\ddagger}$ & \tblval{5.975e-01}
        & \tblval{6.775e-01} & \tblbest{3.876e-01} \\
        Poisson--Boltzmann 2D
        & \tblval{5.723e-01} & \tblbest{4.561e-01}
        & \tblval{4.954e-01} & \tblbest{4.643e-01}
        & \tblval{5.640e-01} & \tblbest{2.297e-01} \\
        Poisson 3D (complex geometry)
        & \tblval{2.872e-01} & \tblbest{2.165e-01}
        & \tblval{4.806e-01} & \tblbest{2.527e-01}
        & \tblval{4.550e-01} & \tblbest{2.393e-01} \\
        Poisson 2D (many-area)
        & \tblval{3.805e-01} & \tblbest{3.312e-01}
        & \tblval{5.632e-01} & \tblbest{3.616e-01}
        & \tblval{8.262e-01} & \tblbest{5.029e-01} \\
        Heat 2D (varying coefficient)
        & \tblval{5.329e-01} & \tblbest{4.856e-01}
        & \tblval{6.093e-01} & \tblbest{4.643e-01}
        & \tblval{6.908e-01} & \tblbest{5.323e-01} \\
        Heat 2D (multiscale)
        & \tblval{4.690e-02} & \tblbest{1.924e-02}
        & \tblval{4.573e-02} & \tblbest{1.717e-02}
        & \tblval{8.771e-02} & \tblbest{1.890e-02} \\
        Heat 2D (complex geometry)
        & \tblval{5.557e-02} & \tblbest{1.519e-02}
        & \tblval{1.794e-02} & \tblbest{1.205e-02}
        & \tblval{2.720e-02} & \tblbest{1.599e-02} \\
        Heat 2D (long time)$^{\dagger}$
        & \tblval{9.990e-01} & \tblval{9.960e-01}
        & \tblval{9.988e-01} & \tblval{9.983e-01}
        & \tblval{9.990e-01} & \tblval{9.981e-01} \\
        Navier--Stokes 2D (lid-driven)
        & \tblval{9.105e-02} & \tblbest{3.406e-02}
        & \tblval{5.190e-02} & \tblbest{3.944e-02}
        & \tblval{5.015e-02} & \tblbest{2.557e-02} \\
        Navier--Stokes 2D (back step)
        & \tblval{8.662e-02} & \tblbest{7.996e-02}
        & \tblval{1.178e-01} & \tblbest{7.593e-02}
        & \tblval{1.029e-01} & \tblbest{6.895e-02} \\
        Navier--Stokes 2D (long time)$^{\dagger}$
        & \tblval{9.948e-01} & \tblval{9.942e-01}
        & \tblval{9.935e-01} & \tblval{9.920e-01}
        & \tblval{9.933e-01} & \tblval{9.945e-01} \\
        Wave 1D
        & \tblval{5.263e-01} & \tblbest{3.603e-01}
        & \tblval{4.729e-01} & \tblbest{3.834e-01}
        & \tblval{4.241e-01} & \tblbest{3.043e-01} \\
        Wave 2D (heterogeneous)
        & \tblval{1.069e+00} & \tblbest{7.819e-01}
        & \tblval{8.419e-01} & \tblbest{7.785e-01}
        & \tblval{8.259e-01} & \tblbest{7.831e-01} \\
        Wave 2D (long time)$^{\dagger}$
        & \tblval{1.000e+00} & \tblval{1.000e+00}
        & \tblval{1.028e+00} & \tblval{1.000e+00}
        & \tblval{1.000e+00} & \tblval{9.999e-01} \\
        Kuramoto--Sivashinsky$^{\dagger}$
        & \tblval{9.744e-01} & \tblval{9.723e-01}
        & \tblval{9.684e-01} & \tblval{9.714e-01}
        & \tblval{9.818e-01} & \tblval{9.744e-01} \\
        Gray--Scott
        & \tblval{9.313e-02} & \tblval{9.304e-02}
        & \tblval{4.417e-01} & \tblbest{2.000e-01}
        & \tblval{9.345e-02} & \tblval{9.338e-02} \\
        Poisson ND
        & \tblval{1.861e-03} & \tblbest{1.384e-03}
        & \tblval{7.980e-04} & \tblbest{6.049e-04}
        & \tblval{4.674e-04} & \tblbest{2.715e-04} \\
        Heat ND
        & \tblval{2.599e-03} & \tblbest{5.898e-04}
        & \tblval{9.063e-04} & \tblbest{2.081e-04}
        & \tblval{5.377e-04} & \tblbest{2.070e-04} \\
        \midrule
        \shortstack[l]{Geometric mean ratio\\(SDT/Vanilla)}
        & \multicolumn{2}{c}{\num{0.718}}
        & \multicolumn{2}{c}{\num{0.717}}
        & \multicolumn{2}{c}{\num{0.672}} \\
        \bottomrule
    \end{tabular}
    }
    \caption{Relative $L_2$ error on all 20 default PINNacle forward problems; lower is better.
    Shading and boldface mark an improvement of at least 5\% within a backbone among the 59 equal-budget comparisons, and none favors Vanilla by that margin.
    The symbol $\dagger$ marks the four problems where both strategies stay above 0.9 L2RE for every backbone: retained for coverage, but not interpreted as comparative evidence.
    The symbol $\ddagger$ marks the one run stopped early (5,000 of 20,000 updates) by numerical instability; that cell is excluded from the summary row and from the 5\% counts.}
    \label{tab:main-results}
\end{table*}

\subsection{Experimental Setup}
\label{sec:experimental-setup}

\paragraph{Benchmarks and metric.}
We use all 20 forward problems in PINNacle~\cite{ZhongkaiHao2024PINNacle}, retaining its operators, domains, conditions, and collocation construction, and report the joint relative $L_2$ error
\begin{equation}
    \operatorname{L2RE}
    = \frac{\bigl(\sum_{n}\left\|\boldsymbol{u}_{\theta}(\boldsymbol{z}_n)-\boldsymbol{u}(\boldsymbol{z}_n)\right\|_2^2\bigr)^{1/2}}
      {\bigl(\sum_{n}\left\|\boldsymbol{u}(\boldsymbol{z}_n)\right\|_2^2\bigr)^{1/2}},
    \label{eq:l2re}
\end{equation}
where lower is better.
Appendix~\ref{app:benchmarks} gives the per-problem protocol.

\paragraph{Architectures and optimization.}
Every target network has width 100, final depth six, and $\tanh$ activations, and the staged configuration uses $K=2$ with $(D_1,D_2)=(3,6)$, so each matched pair of final models has identical parameter counts.
We optimize Eq.~\eqref{eq:pinn-loss} in full batch with AdamW~\cite{Loshchilov2019AdamW} at zero weight decay, matching the Adam update~\cite{Kingma2015Adam} of the default PINNacle implementation, without clipping, warmup, or within-stage decay (Appendix~\ref{app:model-details}).
Both methods receive 20,000 updates and seed 0.
Vanilla ResNet on Poisson2D-Classic became numerically unstable and stopped after 5,000 updates; the primary aggregate excludes this unequal-budget cell and reports the all-suite aggregate as a sensitivity analysis.

\paragraph{Hyperparameter optimization.}
Vanilla and SDT each receive exactly 30 valid multivariate group-TPE trials~\cite{Bergstra2011TPE} per problem--backbone pair in Optuna~\cite{Akiba2019Optuna}: Vanilla searches the learning rate and Xavier initialization scale~\cite{Glorot2010Xavier}, SDT the two initialization scales, later-stage learning-rate ratios, and stage fractions (Appendix~\ref{app:hyperparameters}).
The reported value is each method's minimum final L2RE over its 30 trials on the reporting reference set, an in-sample selection result under a rule applied identically to both arms; Section~\ref{sec:mechanism} also compares the training physics loss, which never uses that solution.

\paragraph{Mechanistic controls.}
A four-arm ablation on Poisson3D with complex geometry (P3D) and lid-driven Navier--Stokes 2D (NS2D) uses the MLP and PirateNet-style backbones.
\emph{E2E-Restart} keeps the full-depth network trainable but rebuilds the optimizer at the same boundaries with the matched SDT run's phase lengths and learning rates, retaining the piecewise schedule without changing depth or transferring a representation.
\emph{Warm-Start (No Freeze)} copies the same shallow prefix into the target-depth network but updates it immediately with the new suffix.
Both inherit that SDT schedule without retuning; all arms share seed 0, the update budget, and the matched Vanilla endpoint.

\subsection{The Curriculum Improves PINNs Across the Benchmark}
\label{sec:main-results}

Table~\ref{tab:main-results} compares the two strategies on all 60 problem--backbone combinations; because vanilla ResNet on Poisson2D-Classic did not complete the matched budget, the primary aggregate uses the remaining 59 equal-budget comparisons.
There, staged training reaches the lower error in 53 of the 58 non-tied comparisons and clears the 5\% margin in 40 of 59; no equal-budget cell favors Vanilla by 5\%, and the largest Vanilla-favoring difference is 2.9\%.
Geometric-mean error is lower by 28.2\% for MLP, 28.3\% for ResNet, and 32.8\% for the PirateNet-style backbone.
Neither aggregation choice drives this: including the unmatched cell gives all-suite reductions of 28.2\%, 27.5\%, and 32.8\%, while restricting to the 16 problems on which at least one method falls below 0.9 L2RE raises them to 33.9\%, 33.0\%, and 39.1\%; that last restriction retains the unmatched cell, and dropping it as well gives 34.3\% for ResNet.

\subsection{The Gain Comes from the Transferred Representation}
\label{sec:mechanism}

E2E-Restart tests whether the gain comes instead from optimizer restarts or the piecewise schedule, and Warm-Start whether copying the prefix suffices without protecting it.

\begin{figure}[!b]
    \centering
    \includegraphics[width=\columnwidth]{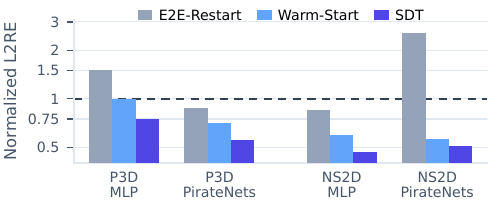}
    \caption{Mechanistic ablation on P3D and NS2D: L2RE normalized by the matched Vanilla result (dashed line, lower is better), logarithmic axis. E2E-Restart has no consistent advantage over Vanilla, and SDT is lower in all four settings.}
    \label{fig:mechanistic-ablation}
\end{figure}

\paragraph{It is not the schedule.}
Figure~\ref{fig:mechanistic-ablation} shows a consistent hierarchy: E2E-Restart improves on Vanilla in two settings and has higher error in the other two, while SDT is lower in every case, so optimizer restarts alone do not match the SDT endpoint.
Warm-Start matches or outperforms Vanilla in all four settings, yet SDT improves on that stronger baseline by 24.6\%, 21.2\%, 22.4\%, and 9.4\%---19.6\% in geometric mean relative to Warm-Start and 43.8\% relative to Vanilla.
Protected transfer is a distinguishing ingredient relative to the restart-only and no-freeze variants.

\paragraph{What crosses the boundary is a representation, not a solution.}
Stage-endpoint diagnostics for 11 matched pairs (Appendix~\ref{app:dynamics}) make the transferred object concrete.
At the first stage boundary the depth-three model can have high solution error---L2RE 1.26 on Heat2D-Multiscale with ResNet, 0.99 on P3D with MLP---yet its prefix supports a final model that beats vanilla training on both.
The finished network improves on its own stage-one endpoint by factors of 1.8 to 73, median 19, so the added depth contributes substantially.

\paragraph{The advantage is visible in the physics objective.}
Because trials are selected on L2RE, we also compare the final unit-weighted training loss of Eq.~\eqref{eq:pinn-loss}, which never sees the reference solution.
SDT attains the lower physics loss in 9 of the same 11 pairs, by factors between 1.2 and 25 (Appendix~\ref{app:physics-loss}), so the L2RE-selected configurations also tend to produce final networks that better satisfy the physics objective.
The two exceptions, P3D with the ResNet and PirateNet-style backbones, end at a higher physics loss but roughly 47\% lower L2RE, consistent with the loss--error decoupling that can arise on complex geometries~\cite{Krishnapriyan2021Failure,Wang2021GradientPathologies}.

\begin{figure}[t]
    \centering
    \includegraphics[width=0.82\columnwidth]{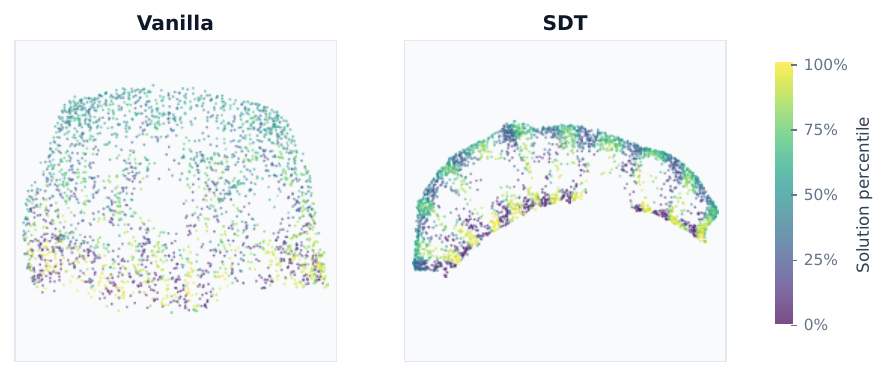}
    \caption{Third-layer ResNet representations on Heat 2D (multiscale): the same 2,400 points under identical t-SNE settings, colored by the percentile of $u(\boldsymbol{z})$. SDT forms a coherent band whereas Vanilla is locally mixed; $V_{20}$ is 33.1\% lower (Appendix~\ref{app:tsne-metric}).}
    \label{fig:representation-tsne}
\end{figure}

\paragraph{The representation itself is better organized.}
Figure~\ref{fig:representation-tsne} inspects the ResNet on Heat 2D (multiscale), where SDT reduces L2RE from \num{4.573e-2} to \num{1.717e-2}.
Post-activation third-layer features at 2,400 evaluation points from both final networks are standardized independently, reduced to 30 PCA dimensions, and embedded with identical t-SNE settings~\cite{vanDerMaaten2008Visualizing}.
Vanilla occupies a broad region in which low, middle, and high solution percentiles repeatedly overlap, whereas SDT produces a continuous band, and a 20-nearest-neighbor measure of local percentile mixing, $V_{20}$, falls from \num{0.320} to \num{0.214}.
Because $V_{20}$ is defined on the embedding it summarizes this visualization; the ablations above provide the primary quantitative diagnostic.

\subsection{The Advantage Survives Perturbing the Configuration}
\label{sec:hyperparameter-stability}

If the gain reflects a better representation rather than a fortunate configuration, it should persist when that configuration is perturbed.
We therefore vary the two hyperparameters shared by both methods, the base learning rate $\eta$ and the Xavier initialization scale $s$, over a $9\times9$ log-spaced grid per method on two randomly selected MLP cells: Heat 2D (complex geometry) and Navier--Stokes 2D (back step).

\begin{figure}[ht]
    \centering
    \includegraphics[width=0.36\textwidth]{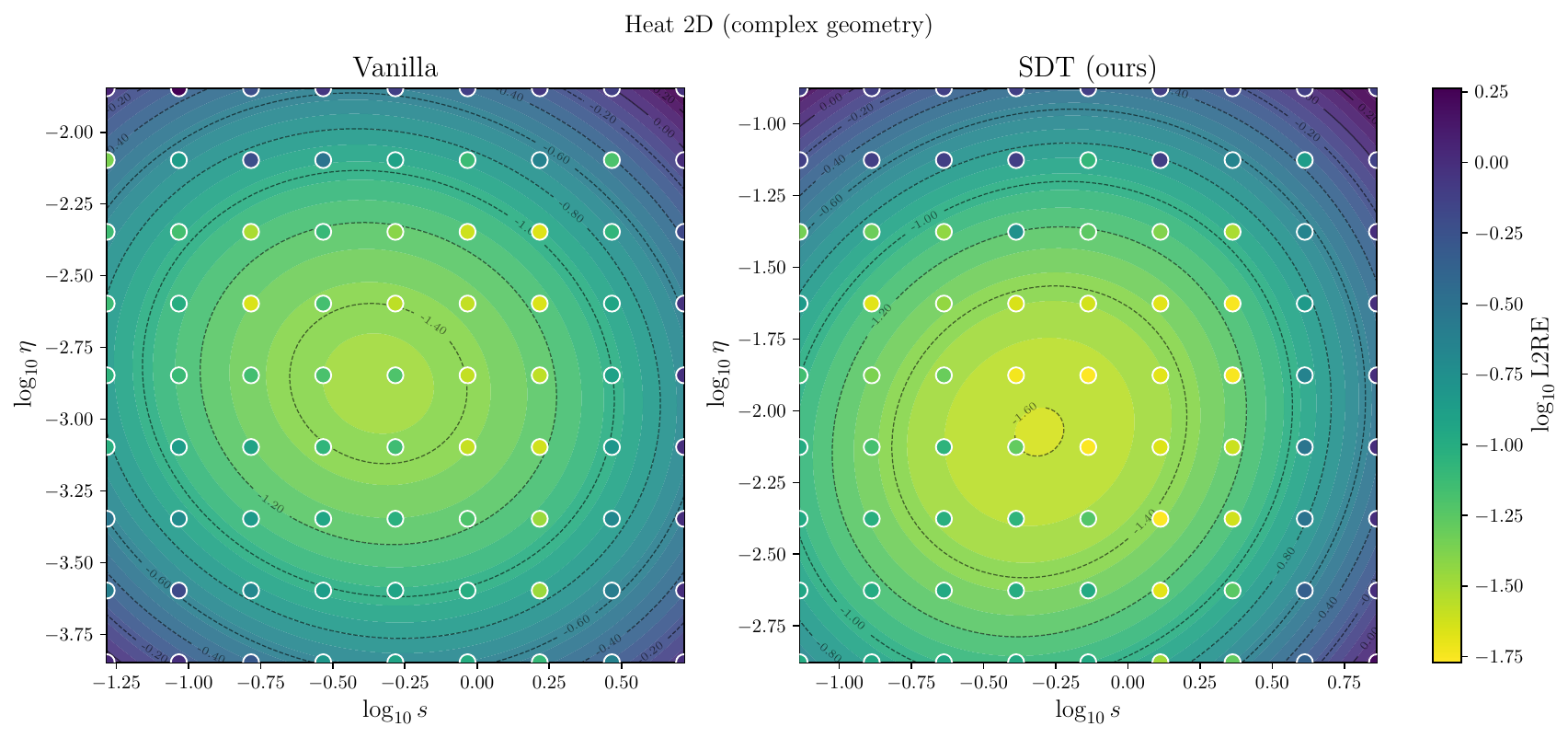}
    \caption{Hyperparameter response surfaces for MLP on Heat 2D (complex geometry): circles mark the evaluated grids, contours are quadratic fits to $\log_{10}\operatorname{L2RE}$. At $\operatorname{L2RE}\leq0.05$ the basin area is 0.673 for Vanilla and 1.203 for SDT.}
    \label{fig:hyperparameter-basin}
\end{figure}

For each grid we fit a quadratic surface $\widehat{\ell}$ to $\log_{10}\operatorname{L2RE}$ and, at a threshold $\tau$ shared by both methods, measure the analytic elliptical area of the basin $\mathcal{B}_{\tau}=\{\widehat{\ell}\leq\log_{10}\tau\}$ in square decades: a larger area means simultaneous changes to $\eta$ and $s$ are less likely to push the model above the target error.
The basin is larger for SDT on both cells, growing from 0.673 to 1.203 at $\tau=0.05$ on Heat 2D (Figure~\ref{fig:hyperparameter-basin}) and from 1.429 to 1.894 at $\tau=0.20$ on Navier--Stokes 2D, factors of 1.79 and 1.33.
On these cells the advantage therefore spans a broader range of configurations, not a narrowly tuned optimum.

\subsection{Progressive Refinement Keeps Paying Off with Depth}
\label{sec:depth-scaling}

Each depth level adds one more representation stage, so a curriculum that works should benefit more, not less, as levels accumulate.
We extend Poisson--Boltzmann 2D to $K\in\{2,\ldots,5\}$ depth levels for the MLP and PirateNet-style backbones, each level adding three hidden layers to a three-layer base, giving depths $n\in\{6,9,12,15\}$, against matched-depth Vanilla models.
Both arms share the depth-three base, which lies outside the staged regime $K\geq2$ and is plotted as their common starting point.

\begin{figure}[t]
    \centering
    \includegraphics[width=0.55\columnwidth]{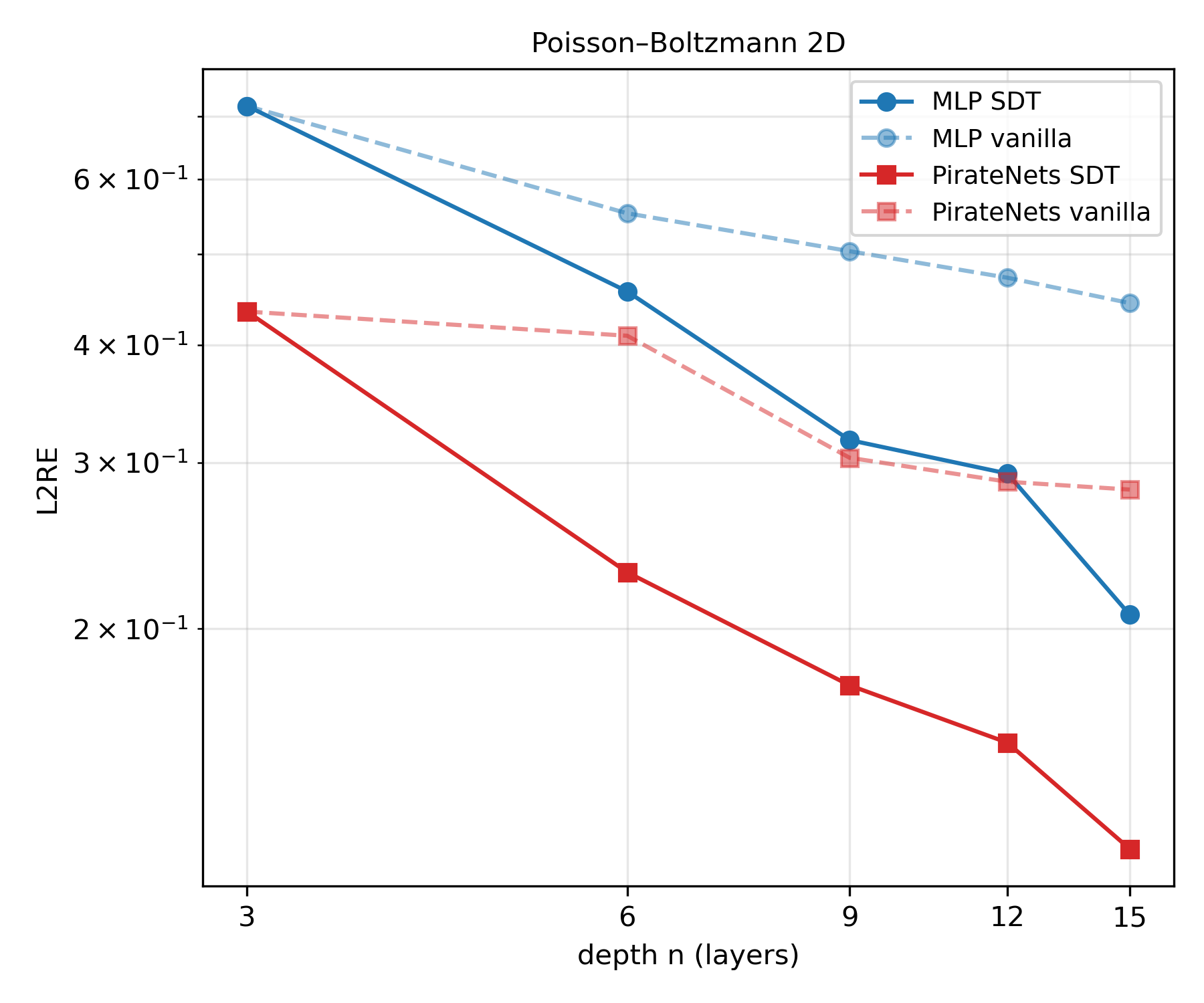}
    \caption{Depth scaling on Poisson--Boltzmann 2D (log--log axes), three layers added per level. The methods coincide at depth three by construction; beyond it SDT attains lower L2RE at every matched depth.}
    \label{fig:depth-scaling}
\end{figure}

Beyond that point every added stage improves accuracy and SDT stays ahead at every matched depth (Figure~\ref{fig:depth-scaling}), reducing L2RE at depth 15 by 53\% for MLP and 58\% for the PirateNet-style backbone.
MLP-SDT matches the Vanilla PirateNet-style network at depths 9--12 and surpasses it at 15 while SDT improves that backbone further, so the two gains are complementary here.
Fitting $\operatorname{L2RE}=cn^{-\gamma}$ by least squares over the five depths, the exponent rises from $0.29$ to $0.74$ for MLP and $0.31$ to $0.78$ for the PirateNet-style backbone: on this problem the curriculum more than doubles the rate at which depth becomes accuracy.

\subsection{Cost and Limitations}
\label{sec:cost}

Instantiating the MAC proxy of Appendix~\ref{app:reproducibility} with the exact stage lengths and PDE dimensions, the reduction over the equal-update comparisons has mean 27.8\% and median 29.6\% but ranges from 2.2\% to 54.9\%, since schedules that spend most of the budget in the final full-depth stage save little; this architecture-level proxy counts dense-layer arithmetic and omits the automatic-differentiation graph.

The empirical claims should be read at the scale of the experiments we run.
Table~\ref{tab:main-results} is a seed-0, fixed-budget comparison over benchmark conditions: each method receives 30 valid trials per cell and we report the minimum final L2RE under its own search space, so the table summarizes selected endpoints rather than run-to-run variance, equivalence under the 5\% display threshold, or held-out performance.
Depth scaling uses one problem and two backbones, basins two cells, and representation geometry one cell.

\FloatBarrier

\section{Conclusion}
\label{sec:conclusion}

PINN hidden representations need not remain incidental by-products of end-to-end solution fitting: SDT gives a concrete recipe for training them as a curriculum---train a shallow prefix through a temporary physics-informed head, discard that head, protect the prefix while adding depth, and fine-tune jointly---while the physics objective, deployed architecture, and inference graph remain fixed.
Across the 20 default PINNacle forward problems and three backbones it obtains lower L2RE in 53 of 58 non-identical equal-budget comparisons and reduces per-backbone geometric-mean error by 28--33\%, and on Poisson--Boltzmann 2D it keeps helping through five depth levels, yielding larger fitted depth-scaling exponents for both backbones.
The four-arm controls and stage-endpoint diagnostics point to protected prefix transfer as a distinguishing ingredient relative to restart-only and warm-start training, and the high-error shallow-stage endpoints support the interpretation that the transferred object is a representation rather than a solution.
Other realizations of the curriculum are possible, but the central lesson is already visible: the representation-learning trajectory is an independent design lever even when the deployed PINN architecture is held fixed.

\FloatBarrier
\bibliography{main}

\clearpage
\appendix
\counterwithin{table}{section}
\counterwithin{figure}{section}

\begin{center}
    {\LARGE\bfseries Technical Appendix}
\end{center}
\vspace{0.5em}

\section{Positioning Relative to Neighboring Methods}
\label{app:positioning}

Several method families stage or grow a network during training, and superficially similar schedules can act on very different objects.
Table~\ref{tab:positioning} focuses on the closest representatives and compares what each method stages and transfers.

\begin{table*}[t]
\centering
{\footnotesize
\setlength{\tabcolsep}{0.85mm}
\begin{tabular}{lllcc}
\toprule
Method & Staged quantity & Carried across stages
  & \shortstack{Depth\\changes} & \shortstack{Same final\\model} \\
\midrule
Greedy layerwise~\cite{Bengio2006GreedyLayerwise}
  & network depth & layer weights & yes & yes \\
Progressive stacking~\cite{Gong2019StackingBERT}
  & Transformer depth & copied blocks & yes & yes \\
Staged Training~\cite{Shen2022StagedTraining}
  & width and depth & loss-preserving growth map & yes & yes \\
RAPTR~\cite{Panigrahi2025RAPTR}
  & active subnetwork & full weights & no & yes \\
LayerLock~\cite{Erdogan2025LayerLock}
  & which layers are trainable & full weights & no & yes \\
\midrule
PirateNets~\cite{SifanWang2024PirateNets}
  & effective block depth & full weights & no & yes \\
PINNup~\cite{Huang2022PINNup}
  & frequency and width & split neurons & no & no \\
Adaptive layerwise~\cite{Krishnanunni2025Layerwise}
  & architecture construction & layer weights & yes & no \\
\midrule
Stacked PINN~\cite{Howard2025Stacked}
  & chained solvers & previous prediction & no & no \\
MGDL / TS-MGDL~\cite{Xu2026MGDL,XuZeng2023MGDL}
  & additive grades & previous residual & no & no \\
\midrule
\textbf{SDT (ours)}
  & hidden-representation depth & frozen PDE-trained prefix & yes & yes \\
\bottomrule
\end{tabular}
}
\caption{Comparison with the closest staged or progressive methods.
``Depth changes'' distinguishes methods that alter the number of layers during training from those that hold a fully instantiated network and vary its loss, its trainable subset, or its inputs.
``Same final model'' records whether training returns a network architecturally identical to an end-to-end baseline of the same target size.}
\label{tab:positioning}
\end{table*}

Stacked and multi-grade methods~\cite{Howard2025Stacked,Xu2026MGDL,XuZeng2023MGDL} carry predictions or residuals into a composite solver, whereas SDT transfers a hidden prefix and returns one serial backbone.
PirateNets~\cite{SifanWang2024PirateNets} keep all layers present and vary their effective influence; SDT instead changes the instantiated depth, so the two mechanisms compose.

\section{Benchmark and Evaluation Details}
\label{app:benchmarks}

\subsection{Official PINNacle Problems}

The main evaluation uses exactly the 20 classes in PINNacle's default forward-problem list~\cite{ZhongkaiHao2024PINNacle}.
We instantiate every class with its default physical parameters, geometry, boundary and initial conditions, and reference solution or reference data.
No parameterized or inverse-problem variants are included in the main table.
Table~\ref{tab:benchmark-sampling} records the input and output dimensions, DeepXDE~\cite{Lu2021DeepXDE} problem type, nominal collocation counts, and L2RE reference for every problem.

\begin{table*}[t]
\centering
{\small
\setlength{\tabcolsep}{1mm}
\begin{tabular}{lrrrrrrrl}
\toprule
Problem & $d_{\mathrm{in}}$ & $d_{\mathrm{out}}$ & Type & $N_r$ & $N_b$ & $N_i$ & $N_{\mathrm{test}}$ & L2RE reference \\
\midrule
Burgers1D & 2 & 1 & TimePDE & 8192 & 2048 & 2048 & 8192 & Reference data \\
Burgers2D & 3 & 2 & TimePDE & 32768 & 8192 & 8192 & 32768 & Reference data \\
\midrule
Poisson2D-Classic & 2 & 1 & PDE & 8192 & 2048 & -- & 8192 & Reference data \\
PoissonBoltzmann2D & 2 & 1 & PDE & 8192 & 2048 & -- & 8192 & Reference data \\
Poisson3D-ComplexGeometry & 3 & 1 & PDE & 32768 & 8192 & -- & 32768 & Reference data \\
Poisson2D-ManyArea & 2 & 1 & PDE & 8192 & 2048 & -- & 8192 & Reference data \\
\midrule
Heat2D-VaryingCoef & 3 & 1 & TimePDE & 32768 & 8192 & 8192 & 32768 & Reference data \\
Heat2D-Multiscale & 3 & 1 & TimePDE & 32768 & 8192 & 8192 & 32768 & Analytic, 20000 points \\
Heat2D-ComplexGeometry & 3 & 1 & TimePDE & 32768 & 8192 & 8192 & 32768 & Reference data \\
Heat2D-LongTime & 3 & 1 & TimePDE & 32768 & 8192 & 8192 & 32768 & Reference data \\
\midrule
NS2D-LidDriven & 2 & 3 & PDE & 8192 & 2048 & -- & 8192 & Reference data \\
NS2D-BackStep & 2 & 3 & PDE & 8192 & 2048 & -- & 8192 & Reference data \\
NS2D-LongTime & 3 & 3 & TimePDE & 32768 & 8192 & 8192 & 32768 & Reference data \\
\midrule
Wave1D & 2 & 1 & PDE & 8192 & 2048 & -- & 8192 & Analytic, 2500 points \\
Wave2D-Heterogeneous & 3 & 1 & PDE & 32768 & 8192 & -- & 32768 & Reference data \\
Wave2D-LongTime & 3 & 1 & TimePDE & 32768 & 8192 & 8192 & 32768 & Analytic, 20000 points \\
\midrule
Kuramoto--Sivashinsky & 2 & 1 & TimePDE & 8192 & 2048 & 2048 & 8192 & Reference data \\
Gray--Scott & 3 & 2 & TimePDE & 32768 & 8192 & 8192 & 32768 & Reference data \\
\midrule
PoissonND & 5 & 1 & PDE & 32768 & 8192 & -- & 32768 & Analytic, 20000 points \\
HeatND & 6 & 1 & TimePDE & 32768 & 8192 & 8192 & 32768 & Analytic, 20000 points \\
\bottomrule
\end{tabular}
}
\caption{Per-problem sampling and evaluation protocol inherited from PINNacle.
The columns $N_r$, $N_b$, $N_i$, and $N_{\mathrm{test}}$ give nominal interior, boundary, initial, and DeepXDE residual-test counts.
The DeepXDE residual-test set is distinct from the reference set used to compute the reported L2RE.}
\label{tab:benchmark-sampling}
\end{table*}

\subsection{Collocation Construction}

PINNacle draws the nominal interior, boundary, and initial candidate points with the Hammersley sequence.
The point sets are constructed once at the beginning of each seeded trial and remain fixed throughout optimization.
Each optimizer update uses the complete collocation set rather than a mini-batch.
DeepXDE may duplicate or repartition nominal boundary candidates when constructing the rows associated with multiple boundary-condition operators, so $N_b$ denotes the benchmark's requested boundary count rather than the sum of all boundary-loss rows after operator expansion.
Wave1D and Wave2D-Heterogeneous are implemented as ordinary PDE data objects rather than TimePDE data objects, and their initial-time constraints are therefore represented through boundary operators instead of a separate $N_i$ pool.

For problems with an analytic solution, the evaluation code uses 2500 uniformly distributed interior points when $d_{\mathrm{in}}=2$ and 20000 points otherwise.
For problems supplied with numerical reference data, it evaluates on every row that contains finite inputs and outputs.
The same seeded reference set is used to rank hyperparameter trials and to report the selected trial in the present experiments.

\section{Architecture and Stage-Transition Details}
\label{app:model-details}

This appendix specifies the $K=2$ instantiation used in the main 60-cell benchmark.

\subsection{Exact Parameter Counts}

Let $d$ and $o$ denote the input and output dimensions, let $w$ denote hidden width, and let $D$ denote final depth.
The MLP and ResNet use the same collection of affine maps, while the ResNet additionally stores a non-trainable input projection.
Their trainable parameter count is
\begin{equation}
    P_{\mathrm{MLP}}=P_{\mathrm{ResNet}}
    = dw+w+(D-1)(w^2+w)+wo.
\end{equation}
The PirateNet-style model contains an input affine map, $D$ adaptive residual blocks with one scalar coefficient each, and the same bias-free output head, giving
\begin{equation}
    P_{\mathrm{Pirate}}
    = dw+w+D(w^2+w+1)+wo.
\end{equation}
For the experimental choice $D=6$ and $w=100$, these reduce to $50600+100(d+o)$ and $60706+100(d+o)$, respectively.
Table~\ref{tab:parameter-counts} groups the official problems by their dimensions and reports the exact final counts.

\begin{table*}[t]
\centering
{\small
\setlength{\tabcolsep}{1mm}
\begin{tabular}{cclrr}
\toprule
$d$ & $o$ & Problems & MLP/ResNet & PirateNet-style \\
\midrule
2 & 1 & Burgers1D; P2D-Classic; PB2D; P2D-ManyArea; Wave1D; KS & 50900 & 61006 \\
3 & 1 & P3D-Complex; four Heat2D; two Wave2D & 51000 & 61106 \\
3 & 2 & Burgers2D; Gray--Scott & 51100 & 61206 \\
2 & 3 & NS2D-LidDriven; NS2D-BackStep & 51100 & 61206 \\
3 & 3 & NS2D-LongTime & 51200 & 61306 \\
5 & 1 & PoissonND & 51200 & 61306 \\
6 & 1 & HeatND & 51300 & 61406 \\
\bottomrule
\end{tabular}
}
\caption{Exact trainable parameter counts of the final depth-six, width-100 models.
Vanilla and staged training have the same count in every row.}
\label{tab:parameter-counts}
\end{table*}

\subsection{Initialization and Transfer}

Every affine weight matrix is sampled with Xavier normal initialization and multiplied by a positive scale selected by hyperparameter optimization.
All affine biases are initialized to zero, and the output head has no bias.
The adaptive residual coefficients remain initialized at zero independently of the affine initialization scale.

Stage 1 initializes the shallow hidden stack and its temporary head with the first-stage scale.
At the expansion boundary, the complete target-depth network and its final head are freshly initialized with the second-stage scale before the learned shallow prefix is copied into the corresponding target layers.
The copy overwrites only the target prefix, so the target suffix and final head retain their fresh second-stage initialization.
The temporary Stage-1 head is never copied into the target model.

For the MLP and ResNet, the copied prefix consists of the first three hidden affine layers.
For the PirateNet-style backbone, it consists of the input affine map and the first three adaptive residual blocks, including their learned $\alpha$ coefficients.
The same prefix is frozen in Stage 2 and unfrozen at the start of Stage 3.

\subsection{Optimizer and Learning-Rate Schedule}

Each stage constructs a fresh AdamW optimizer over exactly the parameters that are trainable in that stage.
The learning rate remains constant within a stage, although the first, second, and third stages may use different optimized values.
The staged step counts are integerized as $T_1=\lfloor f_1T\rfloor$, $T_2=\lfloor f_2T\rfloor$, and $T_3=T-T_1-T_2$, so no optimizer update is lost at a rounding boundary.

Every stage uses AdamW with $(\beta_1,\beta_2)=(0.9,0.999)$, $\epsilon=10^{-8}$, and zero weight decay, which reproduces the Adam update of the default PINNacle implementation.

\section{Hyperparameter Search}
\label{app:hyperparameters}

\subsection{Search Parameterizations}

All searches use Optuna's multivariate group TPE sampler and minimize final L2RE; otherwise unspecified TPE settings are left at their defaults.
All studies share a fixed sampler seed, so the initial random TPE trials are identical across backbones for a given search space.
For every problem--backbone pair, Vanilla and SDT each use exactly 30 valid Optuna trials, so the two arms receive an equal tuning budget as well as an equal number of optimizer updates.
SDT's 30 trials are divided between the untied and tied parameterizations rather than granted to each, and the main table reports the lower error found within that shared budget.
Search bounds were study-specific.
The initial Vanilla search used $s_v\in[0.3,3]$ and $\eta_v\in[3\times10^{-4},3\times10^{-3}]$, and the initial untied SDT search used $\eta_b\in[3\times10^{-4},3\times10^{-2}]$, $r_2\in[0.3,3]$, $r_3\in[0.05,3]$, $s_1,s_2\in[0.3,6]$, $u_1\in[0.1,0.7]$, and $u_2\in[0.05,0.7]$.
For cells requiring broader coverage, both arms were re-searched with expanded bounds under the same rule.
Across all Vanilla searches the sampled space spanned $s_v\in[0.02,20]$ and $\eta_v\in[10^{-5},2\times10^{-2}]$.
The broadest staged space, used for tied MLP on Poisson2D-Classic, was $\eta_b\in[10^{-6},1]$, $s\in[0.05,50]$, $r_{12}\in[0.005,50000]$, $r_3\in[10^{-4},10000]$, and $q\in[0.02,0.49]$.
The vanilla parameterization contains one Xavier initialization scale $s_v$ and one constant learning rate $\eta_v$.
The untied staged parameterization contains a base first-stage learning rate $\eta_b$, two initialization scales, second- and third-stage learning-rate ratios, and two raw stage fractions; it maps to $\eta_1=\eta_b$, $\eta_2=\eta_b r_2$, $\eta_3=\eta_b r_3$, and rescales the first two fractions if needed so that $f_3\ge0.05$.
The tied staged parameterization samples five knobs $(\eta_b,s,r_{12},r_3,q)$ and maps them to $s_1=s_2=s$, $\eta_1=\eta_2=\eta_b r_{12}$, $\eta_3=\eta_b r_3$, and $(f_1,f_2,f_3)=(q,q,1-2q)$.
Learning-rate, ratio, and initialization variables are sampled logarithmically, while stage-fraction variables are sampled linearly.

\subsection{Depth-Scaling Protocol}

The depth-scaling study of Section~\ref{sec:depth-scaling} is a separate set of runs from the main benchmark: both arms were retuned independently at each depth $n\in\{3,6,9,12,15\}$ rather than reusing the depth-six configurations of Table~\ref{tab:main-results}.
The staged runs recover the main-table configuration at $n=6$, so their depth-six points in Figure~\ref{fig:depth-scaling} coincide with the Poisson--Boltzmann 2D entries of that table.
The Vanilla runs do not, and their depth-six points are correspondingly lower.
Figure~\ref{fig:depth-scaling} is therefore matched within itself at every depth, and its Vanilla points should not be read against the main table.

\section{Stage and Objective Diagnostics}
\label{app:dynamics}

We analyze stage endpoints and final objectives for eleven matched Vanilla--SDT pairs spanning four PDEs and all three backbones.
These pairs form a subset of the sixty comparisons in Table~\ref{tab:main-results}; we therefore use them as mechanistic diagnostics rather than as a second full-suite result.

The four-arm controls use the tied SDT schedule so that E2E-Restart and Warm-Start inherit one matched schedule; in two settings this SDT run is weaker than the main-table selection, making the ablation conservative.

\subsection{Stage-One Endpoints and the Use of Depth}

Table~\ref{tab:mechanism-diagnostics} reports the L2RE of the depth-three model at the first stage boundary alongside the final depth-six result.
Two patterns matter for the interpretation of SDT.
First, the stage-one endpoint can have high solution error: its L2RE is 1.256 on Heat2D-Multiscale with ResNet and 0.989 on Poisson3D-ComplexGeometry with MLP.
The prefix extracted from such a model still supports a final network that beats end-to-end training on the same problem, supporting the interpretation that the transferred object is a coordinate representation rather than an approximate solution.
Second, the finished network improves on its own stage-one endpoint by factors between 1.8 and 73, with a median of 19, so the depth added after the first stage contributes substantially to the final result.

\subsection{The Physics Objective}
\label{app:physics-loss}

Because hyperparameter trials are ranked by L2RE, an improvement measured in L2RE alone could in principle reflect the selection rule rather than the optimization.
Table~\ref{tab:mechanism-diagnostics} therefore also compares the final unit-weighted training loss of Eq.~\eqref{eq:pinn-loss}, which is computed from the PDE residual and the boundary and initial conditions and never uses the reference solution.
SDT reaches the lower physics loss in nine of the eleven pairs, by factors between 1.2 and 25.
The two exceptions are Poisson3D-ComplexGeometry with ResNet and with the PirateNet-style backbone, where SDT ends at a higher physics loss yet roughly 47\% lower L2RE.
These cases are consistent with loss--error decoupling that can arise for PINNs on domains with complex geometry, and they limit a purely optimization-based interpretation of SDT on that problem.

\subsection{Sensitivity to the Reported Endpoint}

Both training procedures oscillate between logged evaluations, most strongly on the high-dimensional problems, so we check whether the reported comparison depends on landing at update 20,000.
Replacing each endpoint by the median of the last five logged evaluations moves the geometric-mean ratio over the eleven pairs from \num{0.428} to \num{0.501}: the staged advantage shrinks but persists.
Only two pairs change sign, both on Poisson3D-ComplexGeometry, where the selected schedule devotes as little as 5\% of the budget to the final stage, so a five-point window straddles the last stage boundary and averages away the fine-tuning drop that the schedule is designed to produce.

\begin{table*}[t]
\centering
{\small
\setlength{\tabcolsep}{1mm}
\begin{tabular}{llrrrrrrrr}
\toprule
& & \multicolumn{3}{c}{Final training physics loss}
& \multicolumn{3}{c}{Staged L2RE endpoints}
& \multicolumn{2}{c}{L2RE ratio} \\
\cmidrule(lr){3-5}\cmidrule(lr){6-8}\cmidrule(lr){9-10}
Problem & Backbone
& Vanilla & Staged & ratio
& S1 ($D{=}3$) & final & final/S1
& final & med.\ last 5 \\
\midrule
Heat2D-Multiscale & MLP
  & \num{5.127e-4} & \num{1.029e-4} & \num{0.201}
  & \num{0.368} & \num{1.924e-2} & \num{0.052}
  & \num{0.410} & \num{0.365} \\
Heat2D-Multiscale & ResNet
  & \num{5.014e-4} & \num{6.194e-5} & \num{0.124}
  & \num{1.256} & \num{1.717e-2} & \num{0.014}
  & \num{0.376} & \num{0.401} \\
Heat2D-Multiscale & PirateNet-style
  & \num{7.227e-4} & \num{8.710e-5} & \num{0.121}
  & \num{0.082} & \num{1.890e-2} & \num{0.231}
  & \num{0.215} & \num{0.385} \\
\midrule
NS2D-LidDriven & MLP
  & \num{3.677e-4} & \num{1.490e-5} & \num{0.041}
  & \num{0.731} & \num{3.406e-2} & \num{0.047}
  & \num{0.374} & \num{0.483} \\
NS2D-LidDriven & ResNet
  & \num{4.679e-4} & \num{1.062e-4} & \num{0.227}
  & \num{0.735} & \num{3.944e-2} & \num{0.054}
  & \num{0.760} & \num{0.806} \\
NS2D-LidDriven & PirateNet-style
  & \num{1.455e-4} & \num{1.164e-5} & \num{0.080}
  & \num{0.584} & \num{2.557e-2} & \num{0.044}
  & \num{0.510} & \num{0.550} \\
\midrule
Poisson3D-Complex & MLP
  & \num{2.811e3} & \num{2.042e3} & \num{0.726}
  & \num{0.989} & \num{0.217} & \num{0.219}
  & \num{0.754} & \num{1.099} \\
Poisson3D-Complex & ResNet
  & \num{7.884e0} & \num{5.145e1} & \num{6.526}
  & \num{0.448} & \num{0.253} & \num{0.564}
  & \num{0.526} & \num{0.590} \\
Poisson3D-Complex & PirateNet-style
  & \num{9.634e0} & \num{2.874e2} & \num{29.832}
  & \num{0.547} & \num{0.239} & \num{0.438}
  & \num{0.526} & \num{1.106} \\
\midrule
HeatND & MLP
  & \num{6.189e-4} & \num{1.159e-4} & \num{0.187}
  & \num{7.791e-3} & \num{5.898e-4} & \num{0.076}
  & \num{0.227} & \num{0.296} \\
HeatND & PirateNet-style
  & \num{5.479e-5} & \num{4.450e-5} & \num{0.812}
  & \num{7.350e-3} & \num{2.070e-4} & \num{0.028}
  & \num{0.385} & \num{0.195} \\
\midrule
\multicolumn{2}{l}{Geometric mean}
  & & & \num{0.404}
  & & & \num{0.087}
  & \num{0.428} & \num{0.501} \\
\bottomrule
\end{tabular}
}
\caption{Diagnostics for the eleven matched pairs.
The physics-loss columns use the unit-weighted training loss of Eq.~\eqref{eq:pinn-loss} and are independent of the reference solution; ratios below one favor SDT.
``S1'' is the staged model at the first stage boundary, so ``final/S1'' measures how much the depth added after stage one contributes.
The last two columns compare the reported final-update ratio with a ratio formed from the median of the last five logged evaluations.}
\label{tab:mechanism-diagnostics}
\end{table*}

\subsection{Local Solution-Percentile Variation}
\label{app:tsne-metric}

Figure~\ref{fig:representation-tsne} in the main paper is quantified by a neighborhood measure of how strongly physical states mix locally in the embedding.
Let $\boldsymbol{y}_i\in\mathbb{R}^2$ be the t-SNE coordinate of evaluation point $i$, let $\mathcal{N}_{20}(i)$ contain its 20 Euclidean nearest neighbors in the embedding, and let $c_i\in[0,1]$ be the empirical percentile of the reference solution $u(\boldsymbol{z}_i)$ over the shared evaluation points.
The 20-nearest-neighbor local solution-percentile variation is
\begin{equation}
    V_{20}
    = \frac{
        \displaystyle\sum_{i=1}^{N}\sum_{j\in\mathcal{N}_{20}(i)}
        \lvert c_i-c_j\rvert
    }{
        20N\bigl[Q_{0.95}(c)-Q_{0.05}(c)\bigr]
    },
    \label{eq:local-percentile-variation}
\end{equation}
where $Q_q(c)$ is the $q$-quantile of the sampled percentiles.
Lower $V_{20}$ means neighboring embedded points carry more similar solution values.
Both panels use 2,400 points with exactly 240 per decile, the same percentile assignment, the same color scale, and identical t-SNE settings: cosine distance, perplexity 160, learning rate 100, early exaggeration 12, seed 17, and 2,500 iterations, applied after independent standardization and reduction to 30 PCA dimensions.
The measured values are \num{0.320} for Vanilla and \num{0.214} for SDT.
Because Eq.~\eqref{eq:local-percentile-variation} is evaluated on the two-dimensional embedding, it summarizes the visualization under fixed settings and is not a metric on the representation itself; the ablation of Section~\ref{sec:mechanism} and the physics-loss comparison above carry the quantitative argument.

\section{Compute Accounting and Reproducibility}
\label{app:reproducibility}

\subsection{Dense-Layer Training-MAC Accounting}

The efficiency proxy counts a multiply--accumulate for every dense weight used in a forward pass and approximates backward propagation through the trainable affine maps as twice their forward cost.
It omits bias additions, nonlinearities, optimizer operations, and differential-operator evaluation.
For input dimension $d$, output dimension $o$, width $w$, and depth $D$, the per-point counts are
\begin{align}
    M_D^{\mathrm{MLP}} &= dw+(D-1)w^2+wo, \\
    M_D^{\mathrm{ResNet}} &= 2dw+(D-1)w^2+wo, \\
    M_D^{\mathrm{Pirate}} &= dw+Dw^2+wo.
\end{align}
The extra $dw$ term in the ResNet count is its fixed input projection, while the PirateNet-style backbone counts its learned input map separately from its $D$ residual blocks.
For all three backbones, the trainable Stage-2 suffix and new output head contribute
\begin{equation}
    M_{\mathrm{new}}=3w^2+wo.
\end{equation}
We recover the full-precision stage fractions of each selected configuration and use the same integerization as training:
$T_1=\lfloor f_1T\rfloor$, $T_2=\lfloor f_2T\rfloor$, and $T_3=T-T_1-T_2$.
The per-configuration reduction is
\begin{equation}
    1-\frac{3T_1M_3+T_2(M_6+2M_{\mathrm{new}})+3T_3M_6}{3TM_6}.
\end{equation}
We average this percentage across the matched problem--backbone pairs.
This accounting models the ordinary parameter-gradient cost of the dense layers; it does not attempt to count the PDE-specific higher-order automatic-differentiation graph.

\subsection{Implementation and Randomness}

We implement the networks and optimization in PyTorch 2.10.0, construct the benchmark problems through PINNacle with vendored DeepXDE 1.8.1, and perform hyperparameter optimization with Optuna 4.5.0.
The PINNacle dependency is pinned to commit \texttt{5c3ff7972ca7\allowbreak 0097452cd8ad\allowbreak 8f31c60628d9b49c}.
Experiments used shared GPU nodes equipped with NVIDIA RTX A6000 and Quadro RTX 8000 GPUs.
Before constructing the collocation points and network, we set all pseudorandom seeds to zero.
This keeps the collocation set and the unscaled initialization draw fixed across hyperparameter trials for a given PDE, while the optimized initialization scale changes the magnitude of that draw.

\subsection{Result Selection}

For each problem--backbone pair, Vanilla and SDT each receive 30 valid Optuna trials, and the reported value is the lowest final L2RE found within that method's budget.
Every main-table value is evaluated against a reference solution after the full training budget.
The primary equal-budget aggregate in Section~\ref{sec:main-results} excludes that cell, while Table~\ref{tab:main-results} retains it for full-suite transparency.

\FloatBarrier

\end{document}